\documentclass[dvipsnames]{article} 
\usepackage{iclr2025_conference,times}

\usepackage{hyperref}
\usepackage{url}
\usepackage{graphicx}
\usepackage{booktabs}
\usepackage{mdframed}
\usepackage{enumitem}

\title{Boundless Socratic Learning 
\\with Language Games
}

\author{Tom Schaul\\
Google DeepMind\\
London, UK\\
\texttt{tom@deepmind.com} \\
}

\definecolor{golden}{rgb}{0.746, 0.5625, 0.0078}

\iclrfinalcopy 
\begin{document}

\maketitle

\begin{abstract}
An agent trained within a closed system can master any desired capability, as long as the following three conditions hold: (a) it receives sufficiently informative and aligned feedback, (b) its coverage of experience/data is broad enough, and (c) it has sufficient capacity and resource.
In this position paper, we justify these conditions, and consider what limitations arise from (a) and (b) in closed systems, when assuming that (c) is not a bottleneck.
Considering the special case of agents with matching input and output spaces (namely, language), we argue that such pure recursive self-improvement, dubbed `\emph{Socratic learning},' can boost performance vastly beyond what is present in its initial data or knowledge, and is only limited by time, as well as gradual misalignment concerns. 
Furthermore, we propose a constructive framework to implement it, based on the notion of \emph{language games}.
\end{abstract}

\section{Introduction}

On the path between now and artificial superhuman intelligence \citep[ASI;][]{morris2023levels,grace2024thousands} lies a tipping point, namely when the bulk of a system's improvement in capabilities is driven by \emph{itself} instead of human sources of data, labels, or preferences (which can only scale so far).
As yet, few systems exhibit such \emph{recursive self-improvement}, so now is a prudent time to discuss and characterize what it is, and what it entails.

We focus on one end of the spectrum, the clearest but not the most practical one, 
namely pure self-contained settings of `\emph{Socratic}' learning, closed systems without the option to collect new information from the external world.
We articulate conditions, pitfalls and upper limits, as well as a concrete path towards building such systems, based on the notion of language games.

The central aim of this position paper is to clarify terminology and frame the discussion, with an emphasis on the long run.
It is not to propose new algorithms, nor survey past literature; we pay no heed to near-term feasibility or constraints.
We start with a flexible and general framing, and refine and instantiate these definitions over the course of the paper.

\subsection*{Definitions}

\begin{figure}
    \centering
    \vspace{-2em}
    \includegraphics[width=0.9\linewidth]{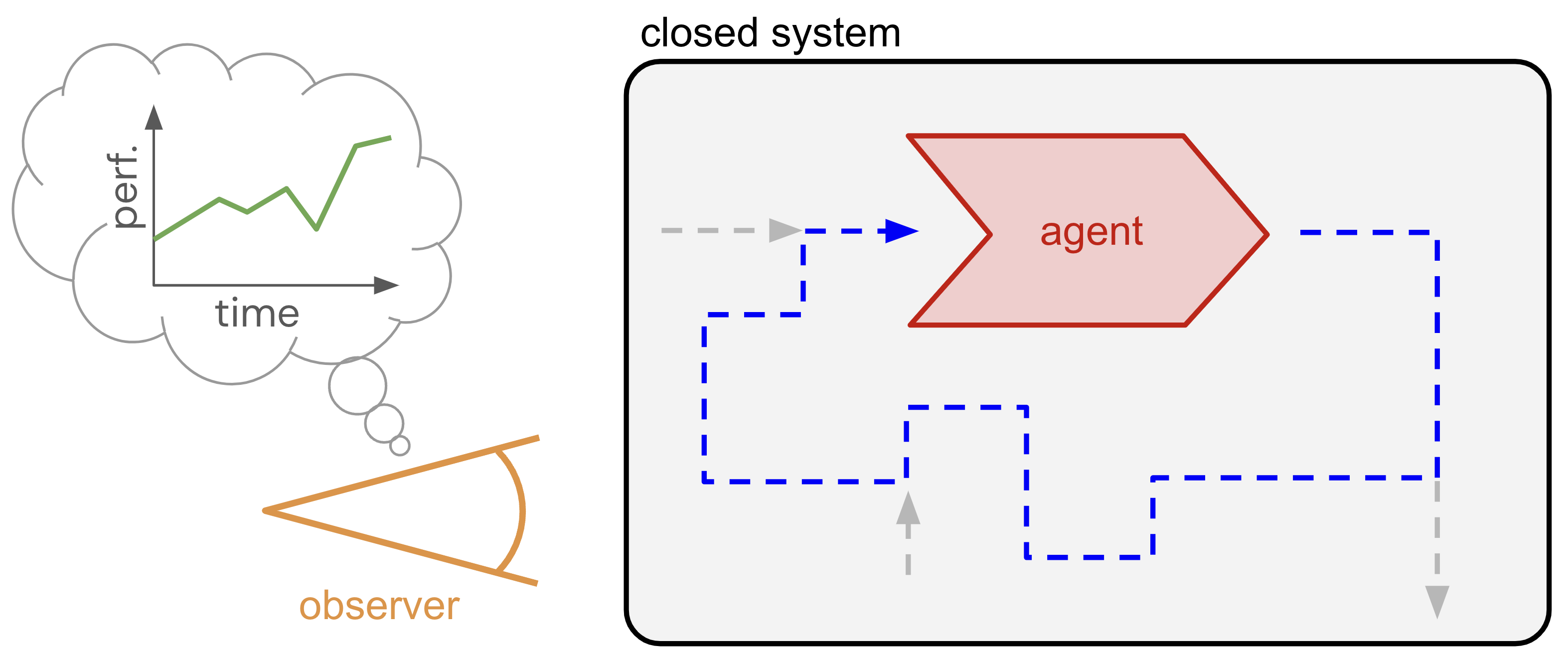}
    \caption{Cartoon depiction of the key definitions. An observer (\textcolor{golden}{\bf gold}) external to a closed system ({\bf black}) assesses the performance (\textcolor{Green}{\bf green}) of an agent (\textcolor{BrickRed}{\bf red}) over time. The process is one of \emph{self-improvement} if agent outputs affect future agent inputs (i.e., some path like \textcolor{NavyBlue}{\bf blue} exists), and performance improves. Further, self-improvement is \emph{recursive} if the agent input and output spaces are compatible, and the process is called `\emph{Socratic learning}' if that space is language. }
    \label{fig:overview}
\end{figure}

Consider a {\bf closed system} (no inputs, no outputs) that evolves over time (see Figure~\ref{fig:overview} for an illustration). 
Within the system is an entity with inputs and outputs, called {\bf agent}, that also changes over time. 
External to the system is an {\bf observer} whose purpose is to assess the {\bf performance} of the agent.
If performance keeps increasing, we call this system-observer pair an {\bf improvement process}.

The dynamics of this process are driven by both the agent and its surrounding system, but setting clear agent boundaries is required to make evaluation well-defined: in fact an agent \emph{is} what can be unambiguously evaluated. 
Similarly, for separation of concerns, the observer is deliberately located outside of the system: As the system is closed, the observer's assessment cannot feed back into the system. Hence, the agent's learning feedback must come from system-internal {\bf proxies} such as losses, reward functions, preference data, or critics.

The simplest type of performance metric is a \emph{scalar} score that can be measured in finite time, that is, on (an aggregation of) episodic tasks. 
Mechanistically, the observer can measure performance in two ways, by \emph{passively} observing the agent's behaviour within the system (if all pertinent tasks occur naturally), or by \emph{copy-and-probe} evaluations where it confronts a cloned copy of the agent with interactive tasks of its choosing.

Without loss of generality, the elements within an agent can be partitioned into three types: \emph{Fixed} elements are unaffected by learning, such as its substrate or unmodifiable code. \emph{Transient} elements do not carry over between episodes, or across to evaluation (e.g., activations, the state of a random number generator). And finally \emph{learned} elements (e.g., weights, parameters, knowledge) change based on a feedback signal, and their evolution maps to performance differences \citep{lu2023reinforcement}.
We can distinguish improvement processes by their implied lifetime; 
some are \emph{open-ended} and keep improving without limit \citep{hughes2024open},
while others converge onto their asymptotic performance after some finite time.%
\footnote{Neither case needs to invoke a notion of optimality \citep{abel2024three}.}

\section{Three Necessary Conditions for Self-improvement}

{\bf Self-improvement} is an improvement process as defined above, but with the additional criterion that the agent's own outputs (actions) influence its future learning. In other words, systems in which agents shape (some of) their own experience stream, potentially enabling unbounded improvement in a closed system.
This setting may look familiar to readers from the reinforcement learning community \citep[RL;][]{sutton2018book}: RL agents' behaviour changes the data distribution it learns on, which in turn affects its behaviour policy, and so on.
Another prototypical instance of a self-improvement process is \emph{self-play}, where the system (often a symmetric game) slots the agent into the roles of both player and opponent, to generate an unlimited experience stream annotated with feedback (who won?) that provides direction for ever-increasing skill-learning.

From its connection to RL, we can derive necessary conditions for self-improvement to work, and help clarify some assumptions about the system. The first two conditions, feedback and coverage, are about feasibility in principle, the third (scale) is about practice.

\subsection{Feedback}
\label{sec:feedback}

Feedback is what gives direction to learning; without it, the process is merely one of self-modification. 
In a closed system where the true purpose resides in the external observer, but can not be accessed directly, feedback can only come from a proxy. 
This creates the fundamental challenge for system-internal feedback is be {\bf aligned} with the observer, and remain aligned throughout the process. It places a significant burden on the system at set-up time, with the most common pitfall being a poorly designed critic or reward function that becomes exploitable over time, resulting in a process that deviates from the observer's intent. 
RL's famed capability for \emph{self-correction} is not applicable here: what can self-correct is behaviour given feedback, but not feedback itself.
Additionally, ideal feedback should be \emph{efficient}, i.e., contain enough information (not too sparse, not too noisy, not too delayed) for learning to be feasible within the time horizon of the system.

\subsection{Coverage}
\label{sec:coverage}

By definition, a self-improving agent determines the distribution of data it learns from. 
To prevent issues like collapse, drift, exploitation or overfitting, it needs to preserve%
\footnote{This may entail conditions on how the system is initialised, as the agent needs to see a first set of inputs before it can produce its own.}
sufficient coverage of the data distribution everywhere the observer cares about.
In most interesting cases, where performance includes a notion of generalisation, that target distribution is not given (the test tasks are withheld), so the system needs to be set up to intrinsically seek coverage, a sub-process classically called \emph{exploration} \citep{ladosz2022exploration}.
Note that aligned feedback is not enough for this on its own: even if a preferred behaviour is never ranked lower than a dispreferred one, that is not tantamount to guaranteeing that the agent will \emph{find} the preferred behaviour.

\subsection{Scale}
\label{sec:capacity}

The research field of RL has produced a lot of detailed knowledge about how to train agents, which algorithms work in which circumstances, an abundance of neat tricks that address practical concerns, as well as theoretical results that characterize convergence, learning dynamics, rates of progress, etc.
It would be futile to try and summarize such a broad body of work here, but see \citet{patterson2023empirical} for a primer.
However, one general observation that matters for our argument is that `RL works at scale': in other words, when scaling up experience and compute sufficiently, even relatively straightforward RL algorithms can solve problems previously thought out of reach \citep[high-profile examples include:][]{tesauro1995temporal,mnih2015human,alphago,alphazero,alphastar,alphaproof-blog}.
For any specific, well-defined practical problem, the details matter (and differ), and greatly impact the efficiency of the learning dynamics; but the asymptotic outcome seems a foregone conclusion. 
The `bitter lesson' of \citet{sutton2019bitter} argues a related point: betting on scaling up computation (as opposed to building in human knowledge) has consistently paid off in the history of AI. 
Hence, with an availability of compute that keeps expanding, the resource constraints of agents (memory and compute) may be a transient concern; not all inefficiencies need to be fixed fully.%
\footnote{Not fully maybe, but learning needs to be efficient \emph{enough} to take advantage of scale without saturating. A specific, timely tension here is around the role of the starting point of learning: some methods that attain mastery while learning purely from scratch (e.g., AlphaZero) while others start with broad competence (LLMs), but may not be as efficient in continuing to learn beyond that.}

\section{Socratic Learning}
The specific type of self-improvement process we consider here is {\bf recursive self-improvement}, where the agent's inputs and outputs are \emph{compatible} (i.e., live in the same space), and outputs become future inputs.%
\footnote{Or at least some of them are fed back. Input and output spaces are not necessarily identical, but they intersect. For example, the agent could be generating code, but perceive natural language, (partly self-generated) code, and execution traces \citep{NEURIPS2023_4b175d84}.
}
This is more restrictive but less mediated than the general case where outputs merely influence the input distribution, most commonly instantiated by a (complex) \emph{environment} that maps agent outputs into inputs.
This type of recursion is an attribute of many open-ended processes, and open-ended improvement is arguably a central feature of ASI \citep[see][]{hughes2024open}.
On the other hand, compatibility is less restrictive than homoiconic self-modification, see Section~\ref{sec:self-referential}.

An excellent example of such a compatible space of inputs and outputs is {\bf language}. 
A vast range of human behaviours are mediated by, and well-expressed,%
\footnote{``Whereof one cannot speak, thereof one must be silent.'' \citep{wittgenstein-tractatus}}
in language, especially in cognitive domains (which are definitionally part of ASI).
As argued by \citet{chalmers2024does} and a few centuries of rationalists before him \citep{cottingham1988rationalists}, language may well be sufficient for thinking and understanding, and not require sensory grounding.
Plus, language has the neat property of being a \emph{soup of abstractions}, encoding many levels of the conceptual hierarchy in a shared space \citep[see also][]{colas2022language}.
A related feature of language is its extendability, i.e., it is possible to develop new languages within an existing one, such as formal mathematics or programming languages that were first developed within natural language.
While special-purpose tools (e.g., interpreters) for these are important for efficiency, natural language may be sufficient as a basis: just like humans can reason `manually' through mathematical expressions when doing mental arithmetic, so can natural language agents \citep{o1-blog}.
And of course, it does not hurt that AI competence on language domains has drastically improved recently, with a lot of momentum since the rise of LLMs.
Early instances of LLM-mediated recursive self-improvement can be glimpsed in the meta-prompts of \cite{fernando2023promptbreeder}, the `action programs' in Voyager's skill library \citet{wang2023voyager},
and most recently, the self-reviewing, paper-generating `AI scientist' \citep{lu2024ai}.

For the remainder of the paper, we will use `{\bf Socratic learning}' to refer to a recursive self-improvement process that operates in language space.
The name is alluding to Socrates' approach of finding or refining knowledge through questioning dialogue and repeated language interactions, 
but, notably, without going out to collect observations in the real world---mirroring our emphasis on the system being closed.
We encourage the reader to imagine an unbroken process of deliberation among a circle of philosophers, maybe starting with Socrates and his disciples, but expanding and continuing undisturbed for millennia: what cultural artifacts, what knowledge, what wisdom could such a process have produced by now?%
\footnote{To make this thought experiment compatible with our setting of a the single agent being evaluated,
assume that the circle maintains the role of spokesperson, whose statements are judged by the observer, and who could be actively queried for evaluation.
}
And then, consider a question that seems paradoxical at first: In principle, how can a closed system produce open-ended improvement?

\vspace{1em}
\begin{mdframed}
\vspace{1mm}
\textsc{Example}
\vspace{-2mm}
\\ \\
\textit{%
To help make these ideas more concrete, we describe a hypothetical but not a priori implausible system \citep[cf.][]{poesia2024learning}.
Consider the domain of mathematical statements (a subset of language).%
\footnote{Note the restriction to a domain like mathematics, with verifiable feedback, is not fully representative of Socratic learning, as is sidesteps most of the challenge of feedback (Section~\ref{sec:feedback}).}
The {\bf observer}'s performance metric is binary: has a proof for the Riemann hypothesis been found?
The {\bf agent} reads and writes mathematical statements and proofs (which are compatible input/output spaces).
The {\bf system} is closed, and contains the agent plus:
\begin{itemize}[topsep=0mm,itemsep=0mm]
    \item a proof verifier (e.g., Lean) 
    \item a collection $C$ of theorems or conjectures.
    \item a {\bf proxy} reward for the agent: $+1$ for each verified new proof of a statement in $C$.
    \item a second collection $L$ of lemmas (or subgoals), initially empty.
\end{itemize}
The system allows the agent to produce proofs, verify them, formulate new statements, and add those to $L$.
Over time, the agent may learn to simplify and decompose existing theorems, accumulate lemmas in $L$, learn to formulate lemmas that are more and more reusable, and increase the fraction of theorems in $C$ for which it can produce valid proofs. It self-improves. At some point, the expanding frontier of verified mathematical knowledge reaches a proof of the Riemann hypothesis, and the observer, satisfied, stops the system.
}
\end{mdframed}

\section{The Fundamental Limits of Socratic Learning}

Among the three necessary conditions for self-improvement, two of them, coverage and feedback apply to Socratic learning \emph{in principle}, and remain irreducible. 
To make their implications as clear as possible, we ignore the third (the scale, practicality and efficiency concerns, see Section~\ref{sec:capacity}) in this section.
We motivate this simplification by taking the long view: if compute and memory keep growing exponentially, scale constraints are but a temporary obstacle. If not, considering a resource-constrained scenario for Socratic learning (akin to studying bounded rationality) may still produce valid high-level insights.

The coverage condition implies that the Socratic learning system must keep generating (language) data, while preserving or expanding diversity over time.
In the LLM age this has come to not seem too far-fetched: We can envision a generative agent initialized with a very broad internet-like distribution that produces a never-ending stream of novel language utterances.
However preventing drift, collapse or just narrowing of the generative distribution in a recursive process may be highly non-trivial \citep{lewis2017deal,shi2024continual}.

The feedback condition requires the system to (a) continue producing feedback about (some subset of) the agent's outputs, which structurally requires a critic that can assess language, and (b) that feedback remains sufficiently aligned with the observer's evaluation metric \citep{christiano2018supervising,bai2022constitutional}.
This is challenging for a number of reasons: Well-defined, grounded metrics in language space are often limited to narrow tasks, while more general-purpose mechanisms like AI-feedback are exploitable, especially so if the input distribution is permitted to shift.
For example, none of the current LLM training paradigms have a feedback mechanism that is sufficient for Socratic learning. Next-token prediction loss is grounded, but insufficiently aligned with downstream usage, and unable to extrapolate beyond the training data. Human preferences are aligned by definition, but prevent learning in a closed system. Caching such preferences into a learned reward model makes it self-contained, but exploitable and potentially misaligned in the long-run, as well as weak on out-of-distribution data.

In other words, pure Socratic learning is possible, but it requires broad data generation with a robust and aligned critic. When those conditions hold, however, the ceiling of its potential improvement is only limited by the amount of resource applied.
Current research has not established successful recipes for this yet, so the next section endeavours to make a concrete but quite general proposal for how to go about it.

\section{Language Games Are All You Need \ldots}

Fortunately, language, learning and grounding are well-studied topics.
A particularly useful concept for us to draw on is Wittgenstein's notion of {\bf language games}.%
\footnote{``I shall also call the whole, consisting of language and the actions into which it is woven, the `language-game'.'' \citep{wittgenstein-philosophical}}
For him, it is not the words that capture meaning, but only the interactive nature of language can do so.
To be concrete here, define a language game as an {\bf interaction protocol} (a set of rules, expressible in code) that specifies the interaction of one or more agents (`players') that have language inputs and language outputs, plus a scalar {\bf scoring function} for each player at the end
of the game.%
\footnote{For simplicity, assume that games are guaranteed to terminate in finite time.}

Language games, thus defined, address the two primary needs of Socratic learning; namely, they provide a scalable mechanism for unbounded interactive data generation and self-play, while automatically providing an accompanying feedback signal (the score).
In fact, they are the logical consequence of the coverage and feedback conditions, almost tautologically so: there is no form of interactive data generation with tractable feedback that is not a language game.%
\footnote{\citet{carse2011finite}'s terminology is handy here too: we mean games of the `finite' type that are played to win, as distinguished from `infinite games' whose aim is to continue playing.}
As a bonus, seeing the process as one of \emph{game-play} immediately lets us import the potential of rich strategic diversity arising from multi-agent dynamics \citep[as spelled out in depth in][]{leibo2019autocurricula,duenez2023social}, which is likely to address at least part of the coverage condition.
It also aligns with our intuition that dynamic, social co-construction (e.g., the circle of philosophers) has an edge over the self-talk of a single person that lives for millennia.
Pragmatically too, games are a great way to get started, given the vast human track record of creating and honing a vast range of games and player skills \citep{berne1968games};
with \citet{nguyen2020games} framing this richness as a demonstration of the fluidity of human agency and (local) motivations.
Derrida might even argue that under the right lens, discourse is already structured as a game.%
\footnote{``Every discourse, even a poetic or oracular sentence, carries with it a system of rules for producing analogous things and thus an outline of methodology.'' \citep{derrida1995points}.
}
\citet{colas2022language} discuss a related set of ideas under the terminology of Vygotskian autotelic agents; while they do not assume a closed system, many of their `internalised social
interactions' could be cast as language games.
A number of common LLM interaction paradigms are also well represented as language games, for example debate \citep{irving2018ai,liang2023encouraging,du2023improving}, role-play \citep{vezhnevets2023generative},
theory of mind \citep{kim2023fantom},
negotiation \citep{lewis2017deal,meta2022human},
jailbreak defense \citep{zeng2024autodefense}, or outside of closed systems, paradigms like RL from human feedback \citep[RLHF,][]{ouyang2022training,bai2022training,gpt4}.

\subsection*{\ldots If You Have Enough of Them \ldots}

Returning to our circle of deliberating philosophers: is there any \emph{one} language game we could imagine them playing for millennia?
Instead, maybe, they are more likely to escape a narrow outcome when playing {\bf many} language games.
It turns out that Wittgenstein (him again) proposed this same idea: he adamantly argued against language having a singular essence or function.%
\footnote{
``But how many kinds of sentence are there? Say assertion,
question, and command?—--There are \emph{countless} kinds: countless different
kinds of use of what we call `symbols,' `words,' `sentences.' And
this multiplicity is not something fixed, given once for all; but new
types of language, new language-games, as we may say, come into
existence, and others become obsolete and get forgotten.'' \citep{wittgenstein-philosophical}, emphasis in original.
}

Using many narrow but well-defined language games instead of a single universal one resolves a key dilemma: For each narrow game, a reliable score function (or critic) can be designed, whereas getting the single universal one right is more elusive \citep[even if possible in principle, as argued by][]{silver2021reward}.%
\footnote{
But, as a prescient Norbert Wiener was warning seven decades ago:
``The machines will do what we ask them to do and not what we ought to ask them to do. [\ldots] We can be humble and live a good life with the aid of the machines, or we can be arrogant and die.'' \citep{wiener}.}
From that lens, the full process of Socratic learning is then a \emph{meta-game}, which schedules the language games that the agent plays and learns from (which is an `infinite' game as per \citet{carse2011finite}).
We posit that in principle, this idea is sufficient to address the issue of coverage (Section~\ref{sec:coverage}). Concretely, if a proxy of the observer's distribution of interest is available (e.g., a validation set of tasks), that can be used to drive exploration in the meta-game.

\subsection*{\ldots And You Play the Right Ones}

Socrates was famously sentenced to death and executed for `corrupting the youth.' We can take this as a hint that a Socratic process is not guaranteed to remain aligned with external observers' intent. Language games as a mechanism do not side-step this either, but they arguably reduce the precision needed: instead of a critic that is aligned at the fine granularity of individual inputs and outputs, all that is needed is a `meta-critic' that can judge which games should be played: maybe no individual language game is perfectly aligned, but what is doable is to filter the many games according to whether they make an overall net-positive contribution (when played and learned about). 
Furthermore, the usefulness of a game does not need to be assessed a priori, but can be judged post-hoc, after playing it for a while. 
Relatedly, a beneficial asymmetry is that it may be much easier to detect deviant emergent behaviour post-hoc than to design games that prevent it. 
All of these properties are forms of structural \emph{leniency} that give the language games framework a vast potential to scale.

Stepping out of our assumption of the closed system for a moment: when we actually build ASI, we will almost surely want to not optimistically trust that alignment is preserved, but instead continually check the process as carefully as possible, and probably intervene and adjust the system throughout training. In that case, explicitly exposing the distribution of games (accompanied by interpretable game descriptions and per-game learning curves) as knobs to the designer may be a useful level of abstraction.

\section{Higher-level Recursions}
So far, we discussed the minimal necessary form of recursion, a form of circularity that feeds (some of) the agent's outputs back to it.
Within the framework of language games, two further types of recursion come to mind. 
The first idea is to tell the agent which game it is playing, and give it the choice to \emph{switch} games, which game to switch to, and when to switch \citep{pislar2021should}.
This is related to hierarchical or goal-conditioned RL, providing the agent with more autonomy and a more abstract action space.
While shifting more responsibility into the agent, this setup could dramatically improve outcomes, as compared to a hardwired game-selection process outside of the agent---but of course this extra freedom could introduce additional risks of collapse or misalignment.

Second, as games are interaction protocols that can be fully represented as code, they can live in a language agent's \emph{output} space. Consequently, the agent could learn to {\bf generate} games for itself to play.%
\footnote{Not strictly just for itself: by defining a language game and communicating its rules to other agents (or itself in a different role) via language, it is possible to produce rich and meaningful multi-agent play. Arguably, the step from being able to set our own goals to the one of communicating rules to co-players was an evolutionary leap for humans \citep{tomasello2022evolution}.}
Initially, it could simply produce local variations of exiting games, which  adapt the difficulty level of theme, later on crafting recombinations of games, and ultimately ending up with \emph{de novo} generation \citep{todd2024gavel}.
This leads to second-order coverage concerns, in the space of language games instead of the space of language, to be addressed with filtering, prioritization, or curricula \citep{jaderberg2019human,parker2022evolving}.

The combination of both of these recursive extensions is an empowered agent that plays the full meta-game of how to improve itself via game generation and play. 
While appealingly elegant, this meta-game lacks the well-defined feedback mechanism of the inner language games, and it is an open research question whether established proxy metrics like learning progress would be sufficient to preserve both the coverage and alignment properties over time.

\subsection*{Self-referential Systems}
\label{sec:self-referential}
The next and final step of recursion is recursive \emph{self-modification}, that is, agents whose actions change their own internals, not merely influencing their input stream.
These methods live on a spectrum characterized by the scope of what can be modified in such a way (and which elements remain fixed), and what amount of introspection, or access to its own workings, is available to the agent \citep{schaul2010metalearning}.
At the extreme end, a \emph{fully self-referential} agent can observe and modify any%
\footnote{Note there always remain some residual frozen bits due to computational primitives, substrate, or---eventually---the laws of physics.
}
aspect of itself, without indirection. In principle, this type of agent has the highest capability ceiling; as asymptotic performance is capped by its fixed structure, unfreezing some of it and making it modifiable can only increase that upper bound---in particular, it is always possible to set the newly-flexible parameters to how they were while frozen, to recover the performance of the less-flexible agent (modulo learning dynamics that could get into the way).
Past proposals for how to design self-referential systems were not (intended to be) practical \citep[e.g.,][]{schmidhuber1993self,schmidhuber2003godel,schmidhuber1997shifting,kirsch2022eliminating}, but modern LLMs' competence in code comprehension and generation is changing the playing field and may soon move these ideas from esoteric to critical.

\section{Conclusion: Open-ended Socratic Learning is Possible}

We set out to investigate how far recursive self-improvement in a closed system can take us on the path to AGI, and are now ready to conclude on an optimistic note.
In principle, the potential of Socratic learning is high, and the challenges we identified (feedback and coverage) are well known. The framework of language games provides a constructive starting point that addresses both, and helps clarify how a practical research agenda could look like. We leave the fleshing out of that roadmap to future work, but the overall direction is becoming apparent.
In particular, an understudied dimension is the breadth and richness of the \emph{many} such language games. We think a great place to start is with processes capable of open-ended game generation.
And not without seeing the irony, we propose all these ideas to scrutiny within an academic setting instead of resorting to self-talk in a closed system. 

\subsubsection*{Acknowledgments} This paper crystallized around a set of seed conversations with Andr\'{e} Barreto and Iulia Com\c{s}a, and was informed by many related discussions, with Wojtek Czarnecki, Diana Borsa, Ed Hughes, Amal Rannen-Triki, Joseph Modayil, Feryal Behbahani, the `1001 Language Games' project team, the DeepMind RL team, among others. Shane Legg, Michael Dennis, Louis Kirsch, Tim Rockt\"{a}schel, David Abel and Chrisantha Fernando provided helpful feedback on an earlier draft.

\bibliography{bib}

\begin{thebibliography}{58}
\providecommand{\natexlab}[1]{#1}
\providecommand{\url}[1]{\texttt{#1}}
\expandafter\ifx\csname urlstyle\endcsname\relax
  \providecommand{\doi}[1]{doi: #1}\else
  \providecommand{\doi}{doi: \begingroup \urlstyle{rm}\Url}\fi

\bibitem[Abel et~al.(2024)Abel, Ho, and Harutyunyan]{abel2024three}
David Abel, Mark~K Ho, and Anna Harutyunyan.
\newblock Three dogmas of reinforcement learning.
\newblock \emph{arXiv preprint arXiv:2407.10583}, 2024.

\bibitem[AlphaProof \& AlphaGeometry(2024)AlphaProof and
  AlphaGeometry]{alphaproof-blog}
AlphaProof and AlphaGeometry.
\newblock {AI} achieves silver-medal standard solving {I}nternational
  {M}athematical {O}lympiad problems.
\newblock \emph{DeepMind blog}, 2024.

\bibitem[Bai et~al.(2022{\natexlab{a}})Bai, Jones, Ndousse, Askell, Chen,
  DasSarma, Drain, Fort, Ganguli, Henighan, et~al.]{bai2022training}
Yuntao Bai, Andy Jones, Kamal Ndousse, Amanda Askell, Anna Chen, Nova DasSarma,
  Dawn Drain, Stanislav Fort, Deep Ganguli, Tom Henighan, et~al.
\newblock Training a helpful and harmless assistant with reinforcement learning
  from human feedback.
\newblock \emph{arXiv preprint arXiv:2204.05862}, 2022{\natexlab{a}}.

\bibitem[Bai et~al.(2022{\natexlab{b}})Bai, Kadavath, Kundu, Askell, Kernion,
  Jones, Chen, Goldie, Mirhoseini, McKinnon, et~al.]{bai2022constitutional}
Yuntao Bai, Saurav Kadavath, Sandipan Kundu, Amanda Askell, Jackson Kernion,
  Andy Jones, Anna Chen, Anna Goldie, Azalia Mirhoseini, Cameron McKinnon,
  et~al.
\newblock Constitutional {AI}: Harmlessness from {AI} feedback.
\newblock \emph{arXiv preprint arXiv:2212.08073}, 2022{\natexlab{b}}.

\bibitem[Berne(1968)]{berne1968games}
Eric Berne.
\newblock \emph{Games people play: The psychology of human relationships},
  volume 2768.
\newblock Penguin Uk, 1968.

\bibitem[Carse(2011)]{carse2011finite}
James Carse.
\newblock \emph{Finite and infinite games}.
\newblock Simon and Schuster, 2011.

\bibitem[Chalmers(2024)]{chalmers2024does}
David~J Chalmers.
\newblock Does thought require sensory grounding? {F}rom pure thinkers to large
  language models.
\newblock \emph{arXiv preprint arXiv:2408.09605}, 2024.

\bibitem[Christiano et~al.(2018)Christiano, Shlegeris, and
  Amodei]{christiano2018supervising}
Paul Christiano, Buck Shlegeris, and Dario Amodei.
\newblock Supervising strong learners by amplifying weak experts.
\newblock \emph{arXiv preprint arXiv:1810.08575}, 2018.

\bibitem[Colas et~al.(2022)Colas, Karch, Moulin-Frier, and
  Oudeyer]{colas2022language}
C{\'e}dric Colas, Tristan Karch, Cl{\'e}ment Moulin-Frier, and Pierre-Yves
  Oudeyer.
\newblock Language and culture internalization for human-like autotelic ai.
\newblock \emph{Nature Machine Intelligence}, 4\penalty0 (12):\penalty0
  1068--1076, 2022.

\bibitem[Cottingham(1988)]{cottingham1988rationalists}
John Cottingham.
\newblock \emph{The Rationalists}.
\newblock Oxford University Press, 1988.

\bibitem[Derrida(1995)]{derrida1995points}
Jacques Derrida.
\newblock \emph{Points… Interviews, 1974-1994}.
\newblock Stanford University Press, 1995.

\bibitem[Du et~al.(2023)Du, Li, Torralba, Tenenbaum, and
  Mordatch]{du2023improving}
Yilun Du, Shuang Li, Antonio Torralba, Joshua~B Tenenbaum, and Igor Mordatch.
\newblock Improving factuality and reasoning in language models through
  multiagent debate.
\newblock \emph{arXiv preprint arXiv:2305.14325}, 2023.

\bibitem[Du{\'e}{\~n}ez-Guzm{\'a}n et~al.(2023)Du{\'e}{\~n}ez-Guzm{\'a}n,
  Sadedin, Wang, McKee, and Leibo]{duenez2023social}
Edgar~A Du{\'e}{\~n}ez-Guzm{\'a}n, Suzanne Sadedin, Jane~X Wang, Kevin~R McKee,
  and Joel~Z Leibo.
\newblock A social path to human-like artificial intelligence.
\newblock \emph{Nature Machine Intelligence}, 5\penalty0 (11):\penalty0
  1181--1188, 2023.

\bibitem[FAIR et~al.(2022)FAIR, Bakhtin, Brown, Dinan, Farina, Flaherty, Fried,
  Goff, Gray, Hu, et~al.]{meta2022human}
Diplomacy~Team FAIR, Anton Bakhtin, Noam Brown, Emily Dinan, Gabriele Farina,
  Colin Flaherty, Daniel Fried, Andrew Goff, Jonathan Gray, Hengyuan Hu, et~al.
\newblock Human-level play in the game of diplomacy by combining language
  models with strategic reasoning.
\newblock \emph{Science}, 378\penalty0 (6624):\penalty0 1067--1074, 2022.

\bibitem[Fernando et~al.(2023)Fernando, Banarse, Michalewski, Osindero, and
  Rockt{\"a}schel]{fernando2023promptbreeder}
Chrisantha Fernando, Dylan Banarse, Henryk Michalewski, Simon Osindero, and Tim
  Rockt{\"a}schel.
\newblock Promptbreeder: Self-referential self-improvement via prompt
  evolution.
\newblock \emph{arXiv preprint arXiv:2309.16797}, 2023.

\bibitem[Grace et~al.(2024)Grace, Stewart, Sandk{\"u}hler, Thomas,
  Weinstein-Raun, and Brauner]{grace2024thousands}
Katja Grace, Harlan Stewart, Julia~Fabienne Sandk{\"u}hler, Stephen Thomas, Ben
  Weinstein-Raun, and Jan Brauner.
\newblock Thousands of ai authors on the future of ai.
\newblock \emph{arXiv preprint arXiv:2401.02843}, 2024.

\bibitem[Hughes et~al.(2024)Hughes, Dennis, Parker-Holder, Behbahani,
  Mavalankar, Shi, Schaul, and Rockt\"{a}schel]{hughes2024open}
Edward Hughes, Michael Dennis, Jack Parker-Holder, Feryal Behbahani, Aditi
  Mavalankar, Yuge Shi, Tom Schaul, and Tim Rockt\"{a}schel.
\newblock Open-endedness is essential for artificial superhuman intelligence.
\newblock \emph{arXiv preprint arXiv:2406.04268}, 2024.

\bibitem[Irving et~al.(2018)Irving, Christiano, and Amodei]{irving2018ai}
Geoffrey Irving, Paul Christiano, and Dario Amodei.
\newblock Ai safety via debate.
\newblock \emph{arXiv preprint arXiv:1805.00899}, 2018.

\bibitem[Jaderberg et~al.(2019)Jaderberg, Czarnecki, Dunning, Marris, Lever,
  Castaneda, Beattie, Rabinowitz, Morcos, Ruderman, et~al.]{jaderberg2019human}
Max Jaderberg, Wojciech~M Czarnecki, Iain Dunning, Luke Marris, Guy Lever,
  Antonio~Garcia Castaneda, Charles Beattie, Neil~C Rabinowitz, Ari~S Morcos,
  Avraham Ruderman, et~al.
\newblock Human-level performance in 3d multiplayer games with population-based
  reinforcement learning.
\newblock \emph{Science}, 364\penalty0 (6443):\penalty0 859--865, 2019.

\bibitem[Kim et~al.(2023)Kim, Sclar, Zhou, Bras, Kim, Choi, and
  Sap]{kim2023fantom}
Hyunwoo Kim, Melanie Sclar, Xuhui Zhou, Ronan~Le Bras, Gunhee Kim, Yejin Choi,
  and Maarten Sap.
\newblock {FANToM}: A benchmark for stress-testing machine theory of mind in
  interactions.
\newblock \emph{arXiv preprint arXiv:2310.15421}, 2023.

\bibitem[Kirsch \& Schmidhuber(2022)Kirsch and
  Schmidhuber]{kirsch2022eliminating}
Louis Kirsch and J{\"u}rgen Schmidhuber.
\newblock Eliminating meta optimization through self-referential meta learning.
\newblock \emph{arXiv preprint arXiv:2212.14392}, 2022.

\bibitem[Ladosz et~al.(2022)Ladosz, Weng, Kim, and Oh]{ladosz2022exploration}
Pawel Ladosz, Lilian Weng, Minwoo Kim, and Hyondong Oh.
\newblock Exploration in deep reinforcement learning: A survey.
\newblock \emph{Information Fusion}, 85:\penalty0 1--22, 2022.

\bibitem[Leibo et~al.(2019)Leibo, Hughes, Lanctot, and
  Graepel]{leibo2019autocurricula}
Joel~Z Leibo, Edward Hughes, Marc Lanctot, and Thore Graepel.
\newblock Autocurricula and the emergence of innovation from social
  interaction: A manifesto for multi-agent intelligence research.
\newblock \emph{arXiv preprint arXiv:1903.00742}, 2019.

\bibitem[Lewis et~al.(2017)Lewis, Yarats, Dauphin, Parikh, and
  Batra]{lewis2017deal}
Mike Lewis, Denis Yarats, Yann~N Dauphin, Devi Parikh, and Dhruv Batra.
\newblock Deal or no deal? end-to-end learning for negotiation dialogues.
\newblock \emph{arXiv preprint arXiv:1706.05125}, 2017.

\bibitem[Liang et~al.(2023)Liang, He, Jiao, Wang, Wang, Wang, Yang, Tu, and
  Shi]{liang2023encouraging}
Tian Liang, Zhiwei He, Wenxiang Jiao, Xing Wang, Yan Wang, Rui Wang, Yujiu
  Yang, Zhaopeng Tu, and Shuming Shi.
\newblock Encouraging divergent thinking in large language models through
  multi-agent debate.
\newblock \emph{arXiv preprint arXiv:2305.19118}, 2023.

\bibitem[Lu et~al.(2024)Lu, Lu, Lange, Foerster, Clune, and Ha]{lu2024ai}
Chris Lu, Cong Lu, Robert~Tjarko Lange, Jakob Foerster, Jeff Clune, and David
  Ha.
\newblock The {AI} scientist: Towards fully automated open-ended scientific
  discovery.
\newblock \emph{arXiv preprint arXiv:2408.06292}, 2024.

\bibitem[Lu et~al.(2023)Lu, Van~Roy, Dwaracherla, Ibrahimi, Osband, Wen,
  et~al.]{lu2023reinforcement}
Xiuyuan Lu, Benjamin Van~Roy, Vikranth Dwaracherla, Morteza Ibrahimi, Ian
  Osband, Zheng Wen, et~al.
\newblock Reinforcement learning, bit by bit.
\newblock \emph{Foundations and Trends in Machine Learning}, 16\penalty0
  (6):\penalty0 733--865, 2023.

\bibitem[Mnih et~al.(2015)Mnih, Kavukcuoglu, Silver, Rusu, Veness, Bellemare,
  Graves, Riedmiller, Fidjeland, Ostrovski, et~al.]{mnih2015human}
Volodymyr Mnih, Koray Kavukcuoglu, David Silver, Andrei~A Rusu, Joel Veness,
  Marc~G Bellemare, Alex Graves, Martin Riedmiller, Andreas~K Fidjeland, Georg
  Ostrovski, et~al.
\newblock Human-level control through deep reinforcement learning.
\newblock \emph{Nature}, 518\penalty0 (7540):\penalty0 529--533, 2015.

\bibitem[Morris et~al.(2023)Morris, Sohl-Dickstein, Fiedel, Warkentin, Dafoe,
  Faust, Farabet, and Legg]{morris2023levels}
Meredith~Ringel Morris, Jascha Sohl-Dickstein, Noah Fiedel, Tris Warkentin,
  Allan Dafoe, Aleksandra Faust, Clement Farabet, and Shane Legg.
\newblock Levels of {AGI}: Operationalizing progress on the path to {AGI}.
\newblock \emph{arXiv preprint arXiv:2311.02462}, 2023.

\bibitem[Nguyen(2020)]{nguyen2020games}
C~Thi Nguyen.
\newblock \emph{Games: Agency as art}.
\newblock Oxford University Press, USA, 2020.

\bibitem[OpenAI et~al.(2023)OpenAI, Achiam, Adler, Agarwal, Ahmad, Akkaya,
  Aleman, Almeida, Altenschmidt, Altman, Anadkat, et~al.]{gpt4}
OpenAI, Josh Achiam, Steven Adler, Sandhini Agarwal, Lama Ahmad, Ilge Akkaya,
  Florencia~Leoni Aleman, Diogo Almeida, Janko Altenschmidt, Sam Altman,
  Shyamal Anadkat, et~al.
\newblock {GPT}-4 technical report.
\newblock \emph{arXiv preprint arXiv:2303.08774}, 2023.

\bibitem[OpenAI et~al.(2024)]{o1-blog}
OpenAI et~al.
\newblock Learning to reason with {LLMs}.
\newblock \emph{OpenAI blog}, 2024.

\bibitem[Ouyang et~al.(2022)Ouyang, Wu, Jiang, Almeida, Wainwright, Mishkin,
  Zhang, Agarwal, Slama, Ray, et~al.]{ouyang2022training}
Long Ouyang, Jeffrey Wu, Xu~Jiang, Diogo Almeida, Carroll Wainwright, Pamela
  Mishkin, Chong Zhang, Sandhini Agarwal, Katarina Slama, Alex Ray, et~al.
\newblock Training language models to follow instructions with human feedback.
\newblock \emph{Advances in neural information processing systems},
  35:\penalty0 27730--27744, 2022.

\bibitem[Parker-Holder et~al.(2022)Parker-Holder, Jiang, Dennis, Samvelyan,
  Foerster, Grefenstette, and Rockt{\"a}schel]{parker2022evolving}
Jack Parker-Holder, Minqi Jiang, Michael Dennis, Mikayel Samvelyan, Jakob
  Foerster, Edward Grefenstette, and Tim Rockt{\"a}schel.
\newblock Evolving curricula with regret-based environment design.
\newblock In \emph{International Conference on Machine Learning}, pp.\
  17473--17498. PMLR, 2022.

\bibitem[Patterson et~al.(2023)Patterson, Neumann, White, and
  White]{patterson2023empirical}
Andrew Patterson, Samuel Neumann, Martha White, and Adam White.
\newblock Empirical design in reinforcement learning.
\newblock \emph{arXiv preprint arXiv:2304.01315}, 2023.

\bibitem[Pislar et~al.(2021)Pislar, Szepesvari, Ostrovski, Borsa, and
  Schaul]{pislar2021should}
Miruna Pislar, David Szepesvari, Georg Ostrovski, Diana Borsa, and Tom Schaul.
\newblock When should agents explore?
\newblock \emph{arXiv preprint arXiv:2108.11811}, 2021.

\bibitem[Poesia et~al.(2024)Poesia, Broman, Haber, and
  Goodman]{poesia2024learning}
Gabriel Poesia, David Broman, Nick Haber, and Noah~D Goodman.
\newblock Learning formal mathematics from intrinsic motivation.
\newblock \emph{arXiv preprint arXiv:2407.00695}, 2024.

\bibitem[Schaul \& Schmidhuber(2010)Schaul and
  Schmidhuber]{schaul2010metalearning}
Tom Schaul and J{\"u}rgen Schmidhuber.
\newblock Metalearning.
\newblock \emph{Scholarpedia}, 5\penalty0 (6):\penalty0 4650, 2010.

\bibitem[Schmidhuber(1993)]{schmidhuber1993self}
J{\"u}rgen Schmidhuber.
\newblock A ‘self-referential’ weight matrix.
\newblock In \emph{ICANN’93: Proceedings of the International Conference on
  Artificial Neural Networks Amsterdam, The Netherlands 13--16 September 1993
  3}, pp.\  446--450. Springer, 1993.

\bibitem[Schmidhuber(2003)]{schmidhuber2003godel}
J{\"u}rgen Schmidhuber.
\newblock G{\"o}del machines: self-referential universal problem solvers making
  provably optimal self-improvements.
\newblock \emph{arXiv preprint cs/0309048}, 2003.

\bibitem[Schmidhuber et~al.(1997)Schmidhuber, Zhao, and
  Wiering]{schmidhuber1997shifting}
J{\"u}rgen Schmidhuber, Jieyu Zhao, and Marco Wiering.
\newblock Shifting inductive bias with success-story algorithm, adaptive
  {L}evin search, and incremental self-improvement.
\newblock \emph{Machine Learning}, 28:\penalty0 105--130, 1997.

\bibitem[Shi et~al.(2024)Shi, Xu, Wang, Qin, Wang, Wang, and
  Wang]{shi2024continual}
Haizhou Shi, Zihao Xu, Hengyi Wang, Weiyi Qin, Wenyuan Wang, Yibin Wang, and
  Hao Wang.
\newblock Continual learning of large language models: A comprehensive survey.
\newblock \emph{arXiv preprint arXiv:2404.16789}, 2024.

\bibitem[Silver et~al.(2016)Silver, Huang, Maddison, Guez, Sifre, Van
  Den~Driessche, Schrittwieser, Antonoglou, Panneershelvam, Lanctot,
  et~al.]{alphago}
David Silver, Aja Huang, Chris~J Maddison, Arthur Guez, Laurent Sifre, George
  Van Den~Driessche, Julian Schrittwieser, Ioannis Antonoglou, Veda
  Panneershelvam, Marc Lanctot, et~al.
\newblock Mastering the game of {Go} with deep neural networks and tree search.
\newblock \emph{nature}, 529\penalty0 (7587):\penalty0 484--489, 2016.

\bibitem[Silver et~al.(2018)Silver, Hubert, Schrittwieser, Antonoglou, Lai,
  Guez, Lanctot, Sifre, Kumaran, Graepel, et~al.]{alphazero}
David Silver, Thomas Hubert, Julian Schrittwieser, Ioannis Antonoglou, Matthew
  Lai, Arthur Guez, Marc Lanctot, Laurent Sifre, Dharshan Kumaran, Thore
  Graepel, et~al.
\newblock A general reinforcement learning algorithm that masters chess, shogi,
  and {Go} through self-play.
\newblock \emph{Science}, 362\penalty0 (6419):\penalty0 1140--1144, 2018.

\bibitem[Silver et~al.(2021)Silver, Singh, Precup, and
  Sutton]{silver2021reward}
David Silver, Satinder Singh, Doina Precup, and Richard~S Sutton.
\newblock Reward is enough.
\newblock \emph{Artificial Intelligence}, 299:\penalty0 103535, 2021.

\bibitem[Sutton(2018)]{sutton2018book}
Richard~S Sutton.
\newblock Reinforcement learning: An introduction.
\newblock \emph{A Bradford Book}, 2018.

\bibitem[Sutton(2019)]{sutton2019bitter}
Richard~S Sutton.
\newblock The bitter lesson.
\newblock \emph{Incomplete Ideas (blog)}, 13\penalty0 (1):\penalty0 38, 2019.

\bibitem[Tesauro et~al.(1995)]{tesauro1995temporal}
Gerald Tesauro et~al.
\newblock Temporal difference learning and td-gammon.
\newblock \emph{Communications of the ACM}, 38\penalty0 (3):\penalty0 58--68,
  1995.

\bibitem[Todd et~al.(2024)Todd, Padula, Stephenson, Piette, Soemers, and
  Togelius]{todd2024gavel}
Graham Todd, Alexander Padula, Matthew Stephenson, {\'E}ric Piette, Dennis~JNJ
  Soemers, and Julian Togelius.
\newblock {GAVEL}: Generating games via evolution and language models.
\newblock \emph{arXiv preprint arXiv:2407.09388}, 2024.

\bibitem[Tomasello(2022)]{tomasello2022evolution}
Michael Tomasello.
\newblock \emph{The evolution of agency: Behavioral organization from lizards
  to humans}.
\newblock MIT Press, 2022.

\bibitem[Vezhnevets et~al.(2023)Vezhnevets, Agapiou, Aharon, Ziv, Matyas,
  Du{\'e}{\~n}ez-Guzm{\'a}n, Cunningham, Osindero, Karmon, and
  Leibo]{vezhnevets2023generative}
Alexander~Sasha Vezhnevets, John~P Agapiou, Avia Aharon, Ron Ziv, Jayd Matyas,
  Edgar~A Du{\'e}{\~n}ez-Guzm{\'a}n, William~A Cunningham, Simon Osindero,
  Danny Karmon, and Joel~Z Leibo.
\newblock Generative agent-based modeling with actions grounded in physical,
  social, or digital space using concordia.
\newblock \emph{arXiv preprint arXiv:2312.03664}, 2023.

\bibitem[Vinyals et~al.(2019)Vinyals, Babuschkin, Czarnecki, Mathieu, Dudzik,
  Chung, Choi, Powell, Ewalds, Georgiev, et~al.]{alphastar}
Oriol Vinyals, Igor Babuschkin, Wojciech~M Czarnecki, Micha{\"e}l Mathieu,
  Andrew Dudzik, Junyoung Chung, David~H Choi, Richard Powell, Timo Ewalds,
  Petko Georgiev, et~al.
\newblock Grandmaster level in {StarCraft II} using multi-agent reinforcement
  learning.
\newblock \emph{Nature}, 575\penalty0 (7782):\penalty0 350--354, 2019.

\bibitem[Wang et~al.(2023)Wang, Xie, Jiang, Mandlekar, Xiao, Zhu, Fan, and
  Anandkumar]{wang2023voyager}
Guanzhi Wang, Yuqi Xie, Yunfan Jiang, Ajay Mandlekar, Chaowei Xiao, Yuke Zhu,
  Linxi Fan, and Anima Anandkumar.
\newblock Voyager: An open-ended embodied agent with large language models.
\newblock \emph{arXiv preprint arXiv:2305.16291}, 2023.

\bibitem[Wiener(1949 / 2013)]{wiener}
Norbert Wiener.
\newblock The machine age / {I}n 1949, he imagined an age of robots.
\newblock \emph{MIT Archives / The New York Times}, D:\penalty0 8, 1949 / 2013.
\newblock URL
  \url{www.nytimes.com/2013/05/21/science/mit-scholars-1949-essay-on-machine-age-is-found.html}.

\bibitem[Wittgenstein(1921)]{wittgenstein-tractatus}
Ludwig Wittgenstein.
\newblock \emph{Tractatus Logico-Philosophicus}.
\newblock 1921.

\bibitem[Wittgenstein(1953)]{wittgenstein-philosophical}
Ludwig Wittgenstein.
\newblock \emph{Philosophical investigations}.
\newblock 1953.

\bibitem[Yang et~al.(2023)Yang, Prabhakar, Narasimhan, and
  Yao]{NEURIPS2023_4b175d84}
John Yang, Akshara Prabhakar, Karthik Narasimhan, and Shunyu Yao.
\newblock Inter{C}ode: Standardizing and benchmarking interactive coding with
  execution feedback.
\newblock In A.~Oh, T.~Naumann, A.~Globerson, K.~Saenko, M.~Hardt, and
  S.~Levine (eds.), \emph{Advances in Neural Information Processing Systems},
  volume~36, pp.\  23826--23854. Curran Associates, Inc., 2023.

\bibitem[Zeng et~al.(2024)Zeng, Wu, Zhang, Wang, and Wu]{zeng2024autodefense}
Yifan Zeng, Yiran Wu, Xiao Zhang, Huazheng Wang, and Qingyun Wu.
\newblock Autodefense: Multi-agent llm defense against jailbreak attacks.
\newblock \emph{arXiv preprint arXiv:2403.04783}, 2024.

\end{thebibliography}
\bibliographystyle{iclr2025_conference}

\end{document}